\documentclass[sigconf,nonacm]{acmart}

\usepackage{tikz}
\usetikzlibrary{shapes,arrows,positioning,calc}
\usepackage{algorithm}
\usepackage{algpseudocode}
\usepackage{tablefootnote}
\usepackage{fvextra}
\RecustomVerbatimEnvironment{verbatim}{Verbatim}{
  breaklines=true,
  breaksymbolleft=,
  breaksymbolright=,
  breakautoindent=true,
  fontsize=\small
}
\AtBeginDocument{%
  }

\begin{document}

% AUTHORS
\author{Jinsong Liu}
\email{jil4047@med.cornell.edu}
% \orcid{0000-0000-0000-0000}
\affiliation{%
  \institution{Dept. of Population Health Sciences, \\ Weill Cornell Medicine, Cornell University}
  \city{New York}
  \state{NY}
  \country{USA}
}

\author{Yuhang Jiang}
\email{yj38@iu.edu}
\affiliation{%
  \institution{
Dept. of Biostatistics and Health Data Science, \\Indiana University School of Medicine}
  \city{Indianapolis}
  \state{IN}
  \country{USA}
}

\author{Ramayya Krishnan}
\email{rk2x@cmu.edu}
\affiliation{%
  \institution{Heinz College of Information Systems and Public Policy, \\Carnegie Mellon University}
  \city{Pittsburgh}
  \state{PA}
  \country{USA}
}

\author{Rema Padman}
\email{rpadman@cmu.edu}
\affiliation{%
  \institution{Heinz College of Information Systems and Public Policy, \\Carnegie Mellon University}
  \city{Pittsburgh}
  \state{PA}
  \country{USA}
}

\author{Yiye Zhang}
\email{yiz2014@med.cornell.edu}
\affiliation{%
  \institution{Dept. of Population Health Sciences, \\ Weill Cornell Medicine, Cornell University}
  \city{New York}
  \state{NY}
  \country{USA}
}

\author{Jiang Bian}
\email{bianji@regenstrief.org}
\affiliation{%
  \institution{Dept. of Biostatistics and Health Data Science, \\Indiana University School of Medicine}
  \city{Indianapolis}
  \state{IN}
  \country{USA}
}

% \renewcommand{\shortauthors}{LastName et al.}

%% TITLE
\title{Closing Reasoning Gaps in Clinical Agents with Differential Reasoning Learning}
% \subtitle{Mining the Cognitive Gap between Agents and Physicians to Automate Clinical Logic Adaptation}

%% ABSTRACT
\begin{abstract}
  Clinical decision support requires not only correct answers but also clinically valid reasoning. We propose Differential Reasoning Learning (DRL), a framework that improves clinical agents by learning from reasoning discrepancies. From reference reasoning rationales (e.g., physician-authored clinical rationale, clinical guidelines, or outputs from more capable models) and the agent's free-form chain-of-thought (CoT), DRL extracts reasoning graphs as directed acyclic graphs (DAGs) and performs a clinically weighted graph edit distance (GED)-based discrepancy analysis. An LLM-as-a-judge aligns semantically equivalent nodes and diagnoses discrepancies between graphs. These graph-level discrepancy diagnostics are converted into natural-language instructions and stored in a Differential Reasoning Knowledge Base (DR-KB). At inference, we retrieve top-$k$ instructions via Retrieval-Augmented Generation (RAG) to augment the agent prompt and patch likely logic gaps. Evaluation on open medical question answering (QA) benchmarks and a Return Visit Admissions (RVA) prediction task from internal clinical data demonstrates gains over baselines, improving both final-answer accuracy and reasoning fidelity. Ablation studies confirm gains from infusing reference reasoning rationales and the top-$k$ retrieval strategy. Clinicians' review of the output provides further assurance of the approach. Together, results suggest that DRL supports more reliable clinical decision-making in complex reasoning scenarios and offers a practical mechanism for deployment under limited token budgets. 
\end{abstract}

%% KEYWORDS

\maketitle

%% INTRODUCTION
\section{Introduction}

Clinical decision support requires not only correct answers but also clinically valid reasoning \citep{chen2024huatuogpt, zuo2025medxpertqa}. In real-world clinical settings, clinicians do not blindly accept a model's output as a standalone label; they pay attention to whether the conclusion is supported by patient-specific evidence, whether key contraindications have been considered, and whether the recommended actions are consistent with established clinical guidelines and standard practice \citep{cabral2024clinical}. As large language models (LLMs) become the backbone of ``clinical agents'' \citep{jin2025agentmd, jiang2025medagentbench, xu2025medagentgym}, the systems that read patient context and generate diagnostic suggestions, treatment recommendations, or risk predictions, the gap between superficially plausible and reliable reasoning becomes a central barrier to real-world deployment and user acceptance. This gap is especially acute in high-stakes settings such as emergency medicine \citep{masanneck2024triage, preiksaitis2024role}, where omission of a single risk factor or an unjustified inference can change disposition decisions, follow-up plans, and ultimately patient outcomes.

\begin{figure*}[htbp]
  \centering
  \includegraphics[width=0.65\textwidth]{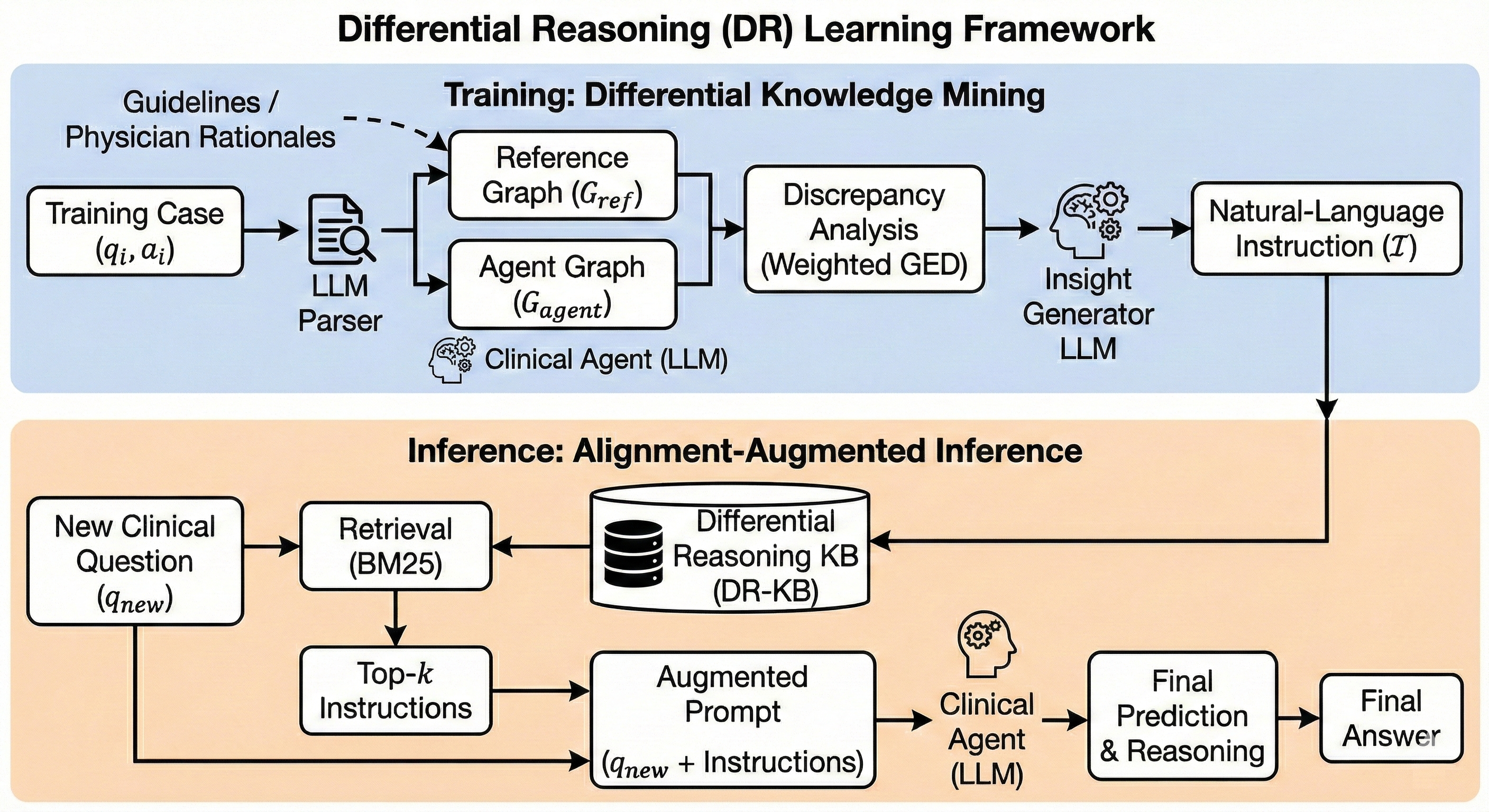}
  \caption{Pipeline for Differential Reasoning Learning (DRL), including Differential Knowledge Mining and Differential Knowledge-Augmented Inference.}
  \label{fig:main_pipeline}
\end{figure*}

A major practical challenge is that contemporary LLM agents typically express reasoning in free-form text (e.g., chain-of-thought, CoT) \citep{wei2022chain}. CoT can be helpful for prompting models to expose intermediate steps, but it is difficult to compare, audit, and improve systematically across cases. The same clinical concept may be paraphrased in countless ways; important evidence may be scattered across a narrative; and many clinically meaningful failures are structural rather than lexical. For example, an agent may reach a correct answer for the wrong reasons, omit an essential intermediate hypothesis, or suggest an inappropriate clinical action even while citing superficially relevant facts. In other cases, the agent may hallucinate irrelevant findings that accidentally steer the conclusion toward the correct option but degrade trust and safety. These issues indicate that focusing only on final-answer accuracy is insufficient: what is needed is a mechanism to detect and correct reasoning gaps, the missing, spurious, or misconnected elements in the agent's reasoning process.

At the same time, clinical domains often provide rich but underutilized sources of high-quality reasoning. These sources include physician-authored clinical rationale, clinical guidelines, expert knowledge, and even the rationales produced by more capable teacher models. Such references encode not just what the answer should be, but why it should be so in a way that is grounded in medical principles. However, turning these references into systematic improvements for an LLM agent remains challenging. Naively adding more exemplars at inference time can be ineffective and quickly runs into context-length limits, while fine-tuning can be costly and may still fail to generalize when the deployment distribution shifts. This motivates a different question: \textbf{Can we learn directly from the discrepancy between an agent's reasoning and an expert's reasoning, and reuse that knowledge to patch future reasoning failures?}

This paper aims to answer the question by casting reasoning alignment as a knowledge discovery problem. Scientific reasoning (and clinical reasoning in particular) is inherently relational: facts support or contradict hypotheses; hypotheses motivate actions; and actions update beliefs or disposition plans. Errors therefore manifest not only as wrong statements but as broken relational structure. We propose to (i) represent reasoning as a structured object, (ii) quantify the difference between agent and reference reasoning, and (iii) distill those differences into reusable guidance or instructions. The result is a framework that turns ``reasoning discrepancies'' into a growing resource for improving scientific and clinical decision-making.

We introduce \textbf{Differential Reasoning Learning (DRL)}, a framework that improves clinical agents by learning from reasoning discrepancies. DRL has two stages: Differential Knowledge Mining (training) and Differential Knowledge-Augmented Inference (testing). During training, the framework compares the agent's reasoning with a reference reasoning trace for the same case and mines the discrepancy signal. During inference, it retrieves the most relevant mined ``patches'' to guide the agent on new cases. The complete procedure is summarized in Algorithm~\ref{alg:drl}.

DRL is built around three design goals that matter in clinical settings. (i) \textbf{Process-level supervision:} the method should target how the answer is obtained, not only whether the final answer is correct. (ii) \textbf{Interpretability and auditability:} the improvement signal should be inspectable by researchers and clinicians, enabling analysis of recurring failure modes. (iii) \textbf{Practicality under domain shift:} the mechanism should help an agent generalize to new tasks and distributions without requiring expensive re-training for every deployment environment.

To make reasoning comparable across cases, we represent it as a \textbf{directed acyclic graph (DAG)} $G=(V,E)$, where nodes $V$ correspond to clinical entities and edges $E$ represent inference steps. Each graph contains nodes for \textbf{facts} (symptoms, labs, history, demographics), \textbf{hypotheses} (diagnoses or clinical conditions, including ruled-out possibilities), and \textbf{actions} (tests, treatments, assessments, monitoring plans). Edges encode clinically meaningful relationships such as a fact supporting a hypothesis, a hypothesis motivating an action, or a fact directly triggering a test \citep{hitchcock2006arguing}. This representation captures the compositional structure of clinical reasoning in a way that is difficult to recover from free-form CoT alone.

We use an LLM as a semantic parser to extract these graphs from unstructured clinical text (Phase~1). For each training case, we construct (i) a \textbf{reference reasoning graph} $G_{ref}$ from (near) gold-standard logic derived from physician-authored clinical rationale, clinical guidelines, expert knowledge, or higher-capacity models, and (ii) an \textbf{agent reasoning graph} $G_{agent}$ extracted from the agent's CoT for the same case. This design addresses a key practical constraint: while physician-authored graphs are costly and difficult to scale for learning, high-quality textual rationales are far more readily available.

Graph extraction also enables a key advantage for knowledge discovery: it makes error analysis compositional. For example, missing an entity such as an important risk factor or a social support related factor can be detected as a missing node, regardless of how it was phrased in the text. Similarly, incorrect clinical actions can be detected as action nodes or edges that should not appear under the case context. This modularity is essential for accumulating reusable knowledge from many diverse cases.

Given $G_{ref}$ and $G_{agent}$, DRL performs a clinically weighted graph edit distance (GED)-based discrepancy analysis (Phase~2) to produce an edit-based diagnostic report. We use the underlying edit structure to capture three clinically meaningful categories of discrepancy: missing clinically relevant factors ($v_{miss}$), hallucinated or irrelevant factors ($v_{halluc}$), and incorrect or incomplete reasoning paths ($e_{diff}$). This diagnostic view directly targets process-level errors: it can identify cases where the agent reaches the correct answer while omitting essential reasoning steps, and it can flag erroneous reasoning that introduces unsupported claims.

Clinical reasoning is rich in paraphrase and contextual equivalence; thus, exact string matching between nodes is insufficient. To make GED robust, we use LLM-as-a-judge to perform semantic node matching and context-based reasonableness checks \cite{li2024llms}. Node types are weighted to reflect clinical importance. The judge outputs an interpretable discrepancy profile: (i) which clinically important nodes are missing or mismatched, (ii) which nodes are hallucinated or irrelevant, and (iii) whether the key reasoning path/edge structure is consistent with the reference. When relevant, we summarize the severity per category (node mismatch, hallucination, wrong reasoning path), but the downstream supervision of instruction generation is driven by the identified edits and affected nodes/edges. The correctness of the answers is also used as a sanity check for the plausibility of the discrepancy analysis. This structured discrepancy signal is crucial for transforming reasoning differences into actionable feedback.

A discrepancy analysis alone does not improve a model; it must be translated into supervision the agent can apply. The core novelty of DRL is to treat graph-level discrepancy diagnostics as opportunities to mine reusable corrective knowledge. An ``Insight Generator'' LLM consumes the discrepancy report (matched/missing nodes, hallucinations, and path/edge errors) together with the case context and produces a structured, actionable \textbf{natural-language instruction} (Phase~3). We generate instructions whenever the analysis surfaces non-trivial missing/hallucinated/path issues. Each instruction is designed to be directly usable in prompts and to generalize beyond the originating case: it specifies what went wrong, why it matters clinically, risks of the error, prevention steps, and contextual triggers (e.g., keywords or patterns that indicate when the guidance applies).

These instructions are stored in a \textbf{Differential Reasoning Knowledge Base (DR-KB)}. DR-KB can be viewed as a repository of distilled ``patches'' that close recurring logic gaps. For example, if the agent misses ``living alone'' as a relevant factor for an elderly fall patient, DRL may store an instruction such as: ``In geriatric fall cases, explicitly assess social support.'' By converting graph-level discrepancies into compact guidance and instructions, DR-KB supports accumulation of corrective knowledge across many cases, a property that mirrors the real-world practice of patient care quality and safety review, where lessons learned from error analysis are codified into checklists and guidelines.

At inference time, DRL uses DR-KB to proactively steer the agent toward clinically valid reasoning (Phase~4). Given a new clinical case or question, we form a retrieval query from the current context, retrieve the top-$k$ most relevant instructions, and inject them into the agent prompt. This is a form of retrieval-augmented generation (RAG), but with a crucial difference: what is retrieved is not raw documents or exemplars, but error-targeted reasoning guidance learned from prior reasoning discrepancies for each case.

This retrieval-based design has several practical advantages. First, it improves the agent without parameter updates, making it easy to deploy and iterate in settings with strict governance or limited resource availability. Second, it is inherently auditable: the retrieved instructions can be inspected, filtered, or edited, and clinical stakeholders can review the system's ``reasoning patches.'' Third, it supports fast adaptation. When new failure modes are observed (e.g., in a new patient population or institutional workflow), DR-KB can be updated by mining additional discrepancies without re-training the base model. Finally, the top-$k$ mechanism provides a direct handle on the trade-off between guidance coverage and context budget, enabling controlled use of external reasoning knowledge.

% \paragraph{Why differential reasoning aligns with AI for science.}
% DR Learning instantiates a general pattern for AI-for-science systems: discover structured discrepancies, distill them into knowledge, and reuse that knowledge to improve future reasoning. Many scientific domains provide supervision in the form of partial rationales, protocols, or expert rationales, yet model reasoning remains difficult to evaluate and align. DR Learning offers a pragmatic bridge. By representing reasoning as graphs and using a discrepancy metric to identify systematic gaps, the framework turns expert feedback into a reusable knowledge base that can guide future predictions. In addition to improving performance, the mined instructions provide an interpretable window into how and why an agent fails, which is essential for scientific validation, safety analysis, and iterative improvement.

We evaluate DRL on both open benchmarks and an internal clinically grounded task. For open-source medical question answering (QA), we use MedQA and MedMCQA, two widely adopted multiple-choice datasets that assess medical knowledge and reasoning. To test robustness under realistic distribution shift, we further curate a real-world hospital dataset focused on \textbf{Return Visit Admission (RVA)} prediction: predicting whether a patient will be readmitted after an emergency department (ED) visit based on the patient's ED assessment. RVA is defined as an ED return visit resulting in hospital admission within 9 days of the index visit, and is clinically meaningful because such patients exhibit higher morbidity and mortality than those admitted initially. We converted clinical notes written by ED physicians into RVA-focused QA pairs, yielding a dataset similar in format to MedQA and MedMCQA.

% Our internal dataset consists of 436 QA pairs: 218 ED visits with RVA and 218 ED visits without RVA for balance. The cases are reviewed by experienced physicians, and physician rationales include index and return impressions/consensus as well as a quality classification indicating whether the return visit is unrelated, illness-related, physician-related, healthcare-system-related, or patient-related. These rationales provide expert context for constructing high-quality reference reasoning graphs $G_{ref}$ and for analyzing reasoning discrepancies in a way that is grounded in clinical practice.

Our experiments demonstrate that learning from reasoning-graph differences yields substantial improvements, especially under domain shift. On the RVA-QA dataset, our DRL framework achieves \textbf{81.28$\pm$0.47} accuracy, outperforming the strongest baseline among the evaluated models (Qwen3-8B at 56.97$\pm$0.57) by \textbf{24.31 points} with low variance across runs (Table~\ref{tab:main_results}). Notably, several baselines that are fine-tuned on open-source medical QA data perform poorly on RVA, highlighting the limits of transferring answer-focused fine-tuning to institution-specific clinical prediction tasks.

This work makes the following contributions:
\begin{itemize}
    \item \textbf{A discrepancy-driven alignment framework.} We propose Differential Reasoning Learning (DRL), which improves clinical agents by explicitly learning from structured discrepancies between agent reasoning and reference reasoning derived from physicians, clinical guidelines, expert knowledge, physician-authored clinical rationale, or higher-capacity models.
    \item \textbf{Graph-based reasoning representation and discrepancy analysis.} We represent clinical reasoning as DAGs and introduce a clinically weighted GED-based discrepancy analysis with an LLM-as-a-judge for semantic matching, yielding interpretable edit diagnostics (missing factors, hallucinations, and path/edge errors) beyond final-answer correctness.
    \item \textbf{Differential Reasoning Knowledge Base (DR-KB).} We convert graph-level discrepancies into reusable natural language instructions and store them in a retrievable knowledge base, enabling accumulation of corrective ``patches'' across future, unseen cases.
    \item \textbf{Retrieval-based inference-time patching.} We present a simple but effective differential knowledge-augmented inference mechanism that retrieves top-$k$ DR-KB instructions and injects them into the agent prompt to close likely logic gaps without parameter updates.
    \item \textbf{Empirical validation on benchmarks and internal EHR data.} We evaluate on MedQA, MedMCQA, and the RVA-QA task, and provide ablations isolating the effects of physician rationales and retrieval depth.
\end{itemize}

The remainder of the paper details the DRL pipeline (reasoning graph extraction, GED-based discrepancy measurement, instruction generation, and retrieval-augmented inference) and evaluates it on open medical QA benchmarks and the internal RVA setting. Overall, our results suggest that learning from reasoning-graph differences offers a practical route toward more reliable clinical agents: it produces explicit, auditable diagnostics of reasoning failures, distills them into reusable knowledge, and applies that knowledge to guide future reasoning in clinically grounded settings. By mining the cognitive gap between agents and experts, we move closer to the goal of trustworthy, reasoning-aligned AI in medicine, ensuring that right clinical decision support is provided for the right reasons. 
% By supporting weaker-capacity models with DRL,  

\section{Related Work}

Aligning clinical agents requires not only accurate predictions but also faithful reasoning processes that can be evaluated, corrected, and improved over time. Our work is most closely related to research in (a) structured representations of clinical reasoning, (b) process-level reasoning supervision and alignment, and (c) agentic memory and error-driven self-improvement. Our work also draws parallels with existing approaches such as instruction tuning and preference learning \citep{ziegler2019fine,ouyang2022training}; while instruction tuning improves task compliance and preference learning optimizes outputs toward human-judged quality, neither provides explicit supervision over reasoning structure, leaving clinical grounding, guideline consistency, and robustness to missing risk factors largely implicit.

\subsection{Structured Representations of Clinical Reasoning}
Recent advances have leveraged Large Language Models (LLMs) as semantic parsers to extract structured representations from unstructured clinical text. While early work focused on entity-centric clinical knowledge graph (KG) construction \citep{wei2025biomedical, wang2025biomedical, wang2025knowledge, wu2025utilizing}, recent systems increasingly emphasize document-specific, dynamically induced structures. For example, \citet{zhang2025medkgent} proposed MedKGent, a multi-agent framework that constructs temporally evolving medical graphs from large-scale biomedical corpora, while \citet{wang2025automated} introduced a retrieval-augmented pipeline that grounds extracted entities and relations in clinical guidelines.

Beyond entity-relation extraction, graph-based representations have been shown to better support reasoning-intensive clinical queries. \citet{wu2025medical} demonstrated that hierarchically indexed graph structures outperform flat retrieval for longitudinal and context-dependent medical reasoning. These approaches motivate the use of structured graphs as an intermediate representation. In contrast to prior work centered on knowledge accumulation or retrieval efficiency, our focus is on extracting reasoning graphs that explicitly encode hypotheses, evidence, and inferential steps, enabling direct comparison between model-generated and reference reasoning processes.

\subsection{Process-Level Reasoning Alignment and Supervision}
Despite strong performance on clinical benchmarks, LLMs often exhibit a mismatch between answer correctness and reasoning validity, a phenomenon commonly referred to as the reasoning gap \cite{wu2025medreason}. Prior studies emphasize that clinical safety depends on alignment with expert reasoning processes rather than outcome accuracy alone \cite{singhal2025toward,nori2023capabilities}. This insight has driven a shift from outcome-supervised reward models to process-supervised approaches \citep{zhang2025lessons}. For instance, \citet{wang2025process} showed that supervising step-by-step reasoning enables reliable detection of latent logical errors that outcome-based supervision overlooks. Complementarily, structural metrics have emerged as principled tools for evaluating reasoning quality. \citet{verma2025grail} used GED to quantify topological discrepancies between generated and expert reasoning structures, effectively penalizing unsupported or hallucinated inference steps. Our work builds on this line of research by using structural discrepancies not only for evaluation but also as an explicit signal to drive downstream correction and learning.

\subsection{Agentic Memory and Error-Driven Self-Improvement}
Learning from past experience is a central theme in agentic systems, where memory mechanisms enable agents to accumulate, retrieve, and reuse knowledge over time \citep{zhang2025survey}. Existing approaches explore diverse memory formats, including textual buffers \citep{packer2023memgpt}, latent embeddings \citep{wang2024memoryllm}, and structured graph memories \citep{chhikara2025mem0,xu2025mem}. Most prior work emphasizes learning from successful trajectories or task completions \citep{tang2025chemagent,zhengsynapse}, with memory serving as a repository of positive experiences.

More recently, researchers have begun to investigate learning from failures. \citet{ouyang2025reasoningbank} proposed ReasoningBank, which curates incorrect reasoning traces to improve future performance. Our approach aligns with this emerging direction but differs in mechanism: rather than storing raw past experiences, we distill reasoning discrepancies into reusable, structured guidance. This formulation connects agentic memory with process-level reasoning alignment, enabling targeted retrieval of corrective instructions conditioned on observed reasoning gaps.

\section{Methodology}

 \textbf{Differential Reasoning Learning (DRL)} improves a clinical agent by learning from reasoning discrepancies between the agent's reasoning and a reference reasoning trace derived from physicians, clinical guidelines, expert rationales, or a more capable teacher model. Because clinical questions are typically multi-factorial involving findings, intermediate hypotheses, and recommended actions, we represent each reasoning trace as a reasoning graph: a directed acyclic graph (DAG) that makes the underlying relational structure explicit. DRL then (i) performs a clinically weighted GED-based discrepancy analysis (with an LLM-as-a-judge for semantic matching) to obtain structured edit diagnostics: what is missing, what is hallucinated, and where the reasoning path deviates, (ii) distills high-discrepancy cases into reusable natural-language ``patches'' (instructions), and (iii) retrieves the most relevant patches at inference time to close likely logic gaps. The overall procedure follows two stages, Differential Knowledge Mining (training) and Differential Knowledge-Augmented Inference (testing), implemented in the four phases below and summarized in Algorithm~\ref{alg:drl}.

\subsection{Phase 1: Reasoning Graph Extraction}

We use a large language model (LLM) as a semantic parser to extract DAG-structured reasoning graphs from unstructured clinical text. For each case, we construct two graphs:
\begin{itemize}
  \item \textbf{Reference reasoning graph ($G_{\mathrm{ref}}$):} a case-specific ``gold-standard'' reasoning trace grounded in clinical guidelines, clinical knowledge, physician-authored clinical rationale, or a higher-capacity model.
  \item \textbf{Agent reasoning graph ($G_{\mathrm{agent}}$):} the reasoning trace induced from the agent's chain-of-thought (CoT) for the same case, extracted by the same graph extractor to avoid underspecified comparisons between free-form rationales.
\end{itemize}

Each graph is denoted as $G=(V,E)$, but we additionally impose a typed, evidence-grounded representation so that reasoning discrepancies can be measured reliably. Because manually curated reasoning graphs by physicians are expensive, we rely on a strong instruction-following LLM to extract $G_{\mathrm{ref}}$ directly from clinical text, incorporating physician-reviewed rationales when available to improve faithfulness. The same schema is used for $G_{\mathrm{agent}}$ so that downstream matching reflects substantive reasoning gaps rather than formatting artifacts. Concretely, the extractor outputs a structured JSON artifact (Appendix~\ref{sec:graph_extraction_prompt}) that partitions $V$ into four components:
\begin{itemize}
  \item \textbf{Facts} $\{F_i\}$ capturing observable findings (e.g., symptoms, vitals, labs, demographics, history) with an optional \texttt{value} field and a \texttt{polarity} in \{\texttt{present}, \texttt{absent}, \texttt{uncertain}\};
  \item \textbf{Hypotheses} $\{H_j\}$ capturing diagnoses or clinical conditions with a level of \texttt{confidence} in \{\texttt{high}, \texttt{medium}, \texttt{low}, \texttt{uncertain}\};
  \item \textbf{Actions} $\{A_k\}$ capturing clinical actions with \texttt{action\_type} in \{\texttt{TEST}, \texttt{TREAT}, \texttt{ASSESS}, \texttt{OBSERVE}, \texttt{PRESCRIBE}\} and a \texttt{status} field (e.g., \texttt{recommended}, \texttt{contraindicated});
  \item \textbf{Final} node representing the QA answer or final disposition or diagnosis.
\end{itemize}
Each node carries an evidence list \texttt{support} containing verbatim quotes from the case text and (when applicable) references to other node IDs, enabling both manual inspection and LLM-based adjudication.

Edges $E$ are directed, binary relations with a single source and a single destination, encoded as tuples \texttt{(src, dst, type)} where \texttt{type} $\in$ \{\texttt{supports}, \texttt{contradicts} (i.e., rules out), \texttt{suggests\_test}\}. The extractor instantiates edges following common clinical reasoning patterns, including Fact$\rightarrow$Hypothesis, Hypothesis$\rightarrow$Action, Fact$\rightarrow$Action, Action$\rightarrow$Hypothesis, and links into the Final node. Each edge additionally stores a short \texttt{justification} field in natural language, which we retain as auxiliary metadata for interpretability. We constrain the overall structure to a DAG to mirror the forward flow from evidence to intermediate hypotheses to actions and conclusions, and to keep downstream graph matching well-behaved. The detailed JSON schema of the graphs is provided in Appendix~\ref{sec:graph_extraction_prompt}.

\subsection{Phase 2: Measuring the Reasoning Gap}
We quantify reasoning alignment using a clinically weighted GED-based discrepancy analysis, defined via the minimum-cost edit script that aligns $G_{\mathrm{agent}}$ to $G_{\mathrm{ref}}$. For interpretability, we explicitly separate the edits into clinically relevant missing factors, hallucinated factors, and structural (edge/path) inconsistencies, and we treat these components as the primary output:
\begin{equation}
  \begin{aligned}
  \mathbf{D}(G_{\mathrm{agent}}, G_{\mathrm{ref}}) & =
  \big(d_{\mathrm{miss}}, d_{\mathrm{halluc}}, d_{\mathrm{path}}\big) \\
  & =
  \left(\sum c(v_{\mathrm{miss}}),\ \sum c(v_{\mathrm{halluc}}),\ \sum c(e_{\mathrm{diff}})\right).
  \end{aligned}
\end{equation}
Here $v_{\mathrm{miss}}$ are clinically important nodes present in $G_{\mathrm{ref}}$ but absent in $G_{\mathrm{agent}}$, $v_{\mathrm{halluc}}$ are nodes introduced by the agent that are unsupported or irrelevant in context, and $e_{\mathrm{diff}}$ captures edge-level discrepancies (e.g., incorrect directional support, missing links required for the reference reasoning path, or spurious inference steps). DRL conditions instruction generation on the discrepancy report (the categorized edits and affected nodes/edges), which is stable across tasks and prompts.

To make GED robust to clinical paraphrases, we use an LLM-as-a-judge to perform semantic node matching and context-sensitive reasonableness checks, rather than relying on exact string overlap. We assign higher penalties to failures on clinically consequential node types, weighting Facts by $1.0\times$, Hypotheses by $1.5\times$, and Actions by $2.0\times$. The judge outputs a discrepancy profile over three categories: missing/mismatched nodes, hallucinated/irrelevant nodes, and reasoning path/edge errors. When helpful for summarization, the judge also reports a normalized severity per category. Answer correctness is used only as a sanity check for plausibility of the discrepancy analysis (e.g., to flag inconsistent evaluations), not to impose fixed numeric regimes. The full prompt for discrepancy analysis and instruction generation is provided in Appendix~\ref{sec:ged_instruction_prompt} (Step~1).

\subsection{Phase 3: Differential Reasoning Knowledge Base}

Rather than treating discrepancies as isolated failures, DRL converts discrepancies into reusable guidance. An Insight Generator LLM consumes the GED diagnostics including matched and missing nodes, hallucinations, and path errors, and produces a structured, actionable instruction that specifies (i) the error type, (ii) the clinical context in which it arises, and (iii) concrete prevention or remediation steps (including risks, contraindications, trigger keywords, and the underlying clinical principle):
\begin{displaymath}
    \mathcal{I} = \text{LLM}(\text{``Graph difference } Diff(G_{\mathrm{agent}}, G_{\mathrm{ref}}) \text{ in context } C\text{''})
\end{displaymath}
We store the resulting instructions in the \textbf{Differential Reasoning Knowledge Base (DR-KB)} for later retrieval. Appendix~\ref{sec:ged_instruction_prompt} (Step~2) shows the prompt used for instruction generation.

\textit{Example:} If the agent fails to account for \emph{living alone} when reasoning about disposition after a geriatric fall, the generated patch may be: \textit{``In geriatric fall cases, explicitly assess social support (e.g., living situation) before determining discharge safety.''}

\subsection{Phase 4: Differential Knowledge-Augmented Inference}

At inference, we operationalize DR-KB as an error-targeted retrieval module. We index instructions with BM25 \citep{robertson2009probabilistic} for lexical retrieval and, given a new clinical question (and its associated context), we:
\begin{enumerate}
    \item form a BM25 query from the current clinical context,
    \item retrieve the top-$k$ most relevant DR-KB instructions, and
    \item inject the retrieved instructions into the agent prompt to dynamically patch likely reasoning gaps.
\end{enumerate}

This is a retrieval-augmented generation (RAG) mechanism in which the retrieved items are not raw documents or exemplars, but discrepancy-derived reasoning patches mined from prior failures.

\begin{algorithm}[t]
\caption{DR Learning: Training and Inference Process}
\label{alg:drl}
\begin{algorithmic}[1]
\State \textbf{Input:} Training set $\mathcal{D} = \{(q_i, a_i)\}_{i=1}^N$
\State \textbf{Initialize:} DR-KB $\mathcal{K} \leftarrow \emptyset$
\State \textbf{Stage 1: Differential Knowledge Mining}
\For{each case $(q_i, a_i)$ in $\mathcal{D}$}
    \State $G_{\mathrm{ref}} \leftarrow \text{ExtractGraph}(q_i, a_i, \text{physician rationales})$
    \State $\hat{a}_i, \text{CoT}_i \leftarrow \text{Agent}(q_i)$
    \State $G_{\mathrm{agent}} \leftarrow \text{ExtractGraph}(\text{CoT}_i)$
    % \State $\Delta \leftarrow \text{GED}(G_{\mathrm{ref}}, G_{\mathrm{agent}})$
    % \If{$\Delta > \text{Threshold } \tau$}
    %     \State Instruction $\mathcal{I} \leftarrow \text{GenerateInsight}(G_{\mathrm{ref}}, G_{\mathrm{agent}}, q_i)$
    %     \State $\mathcal{K}.\text{add}(\mathcal{I})$
    % \EndIf
    \State $\mathbf{D}_i \leftarrow \text{AnalyzeDiscrepancy}{\mathrm{GED}}(G_{\mathrm{ref}}, G_{\mathrm{agent}})$ \Comment{semantic matches + missing/hallucinated/path edits}
    \If{$\mathbf{D}_i \neq \emptyset$}
    \State Instruction $\mathcal{I} \leftarrow \text{GenerateInsight}(G_{\mathrm{ref}}, G_{\mathrm{agent}}, \mathbf{D}_i, q_i)$
    \State $\mathcal{K}.\text{add}(\mathcal{I})$
  \EndIf
\EndFor
\State \textbf{Stage 2: Differential Knowledge-Augmented Inference}
\For{new clinical question $q_{\mathrm{new}}$}
    \State $\mathcal{I}_{\mathrm{retrieved}} \leftarrow \text{Retrieve}(\mathcal{K}, q_{\mathrm{new}}, k)$
    \State Prompt $\leftarrow$ ``Question: $q_{\mathrm{new}}$. Guidelines: $\mathcal{I}_{\mathrm{retrieved}}$''
    \State Prediction $\leftarrow \text{Agent}(\text{Prompt})$
\EndFor
\end{algorithmic}
\end{algorithm}

\section{Experiments}

\subsection{Experimental Setup}
We conduct comprehensive experiments to evaluate DRL on open-source medical QA benchmarks and an internal, clinically grounded dataset. For public benchmarks, we use MedQA \citep{jin2021disease} and MedMCQA \citep{pal2022medmcqa}, which are widely used to assess medical QA and reasoning \citep{wu2025medreason,chen2024huatuogpt,yun2025med}. We additionally curate a dataset for Return Visit Admission (RVA) prediction \citep{zhang2024aspiring} from electronic health records (EHR) at a large, urban ED by converting ED clinical notes into RVA-focused QA pairs. The two sets of experiments, on MedQA/MedMCQA and on RVA, are meant to test DRL's ability on different tasks. The former tests reasoning for facts and diagnostic tasks, whereas the latter tests reasoning for prognosis and predictive tasks. Each task has only one right answer. 

Across all experiments, we use Qwen3-8B (Qwen) \citep{yang2025qwen3} and LLaMA-3.1-8B-Instruct (LLaMA) \citep{dubey2024LLaMA} as the backbone agent models due to their strong performance and practical deployment size, and to show the robustness of DRL across different base models. For reasoning-graph extraction and GED-based discrepancy analysis, we use Gemini-2.5-Flash \citep{comanici2025gemini}. This choice is driven by our requirement to run the EHR dataset in a HIPAA-compliant environment where only HIPAA-compliant Gemini models are available; for consistency, we use Gemini-2.5-Flash across all datasets. For converting internal EHR data to the QA format, we use Gemini-2.5-Pro \citep{comanici2025gemini}.

We compare DRL against strong medical QA agent baselines: MedReason-8B \citep{wu2025medreason}, fine-tuned on synthetic CoT data derived from MedQA and MedMCQA; MedPRM-8B \cite{yun2025med}, also trained on these benchmarks; and HuatuoGPT-o1-8B \citep{chen2024huatuogpt}, a state-of-the-art medical reasoning model fine-tuned on multiple medical QA datasets, including MedQA and MedMCQA.

To account for the stochasticity of LLM generation, we repeat each experiment five times and report mean accuracy with standard deviation. Three expert clinicians reviewed the reasoning output for validation.

\subsection{Datasets}

\subsubsection{Open Medical QA Datasets}
We evaluate on open-source medical QA benchmarks MedQA \citep{jin2021disease} and MedMCQA \citep{pal2022medmcqa}, which consist of multiple-choice questions covering clinical knowledge and reasoning. 
As both datasets consist of diverse medical topics, we restrict the evaluation to questions related to sepsis, chest pain, and stroke by keyword matching to align with the clinical scope of our internal EHR dataset. For each dataset, we randomly sample 500 training examples and 100 test examples, ensuring a controlled experimental setting that aligns the clinical scope and data scale across benchmarks. Accuracy serves as the primary metric.

\subsubsection{Return Visit Admission Prediction}
Return visit admissions (RVA) are ED return visits resulting in hospital admission within 9 days of the index visit \citep{zhang2024aspiring}. These patients exhibit higher morbidity and mortality than those admitted initially. Drivers include rapid disease progression, patient non-adherence, and physician error. We therefore convert ED notes into RVA-focused QA pairs, framing a binary prediction task that asks whether a patient will experience an RVA based on the summarized ED encounter, yielding a dataset comparable in format to open QA benchmarks such as MedQA and MedMCQA.

We curated 218 ED visits with RVA from electronic health records (EHR), further reviewed by experienced ED physicians who provided reasoning rationales. Reasoning rationales include initial and return diagnostic impressions, and classification for readmission reasons. The reasoning indicates whether the return visit is illness related, physician related, healthcare system related, patient related, or unpreventable. These rationales provide expert knowledge for constructing high-quality reference reasoning graphs $G_{ref}$. For a balanced dataset, we randomly sampled 218 ED visits without RVA, totaling 436 QA pairs. To ensure conversion quality, we use the most capable model available at the study site (Gemini-2.5-Pro) to generate initial QA pairs. The prompt for this conversion is provided in Appendix~\ref{sec:clinical_note_to_qa_prompt}.

\subsection{Experimental Results}
\begin{figure}[htbp]
  \centering
  \includegraphics[width=0.9\linewidth]{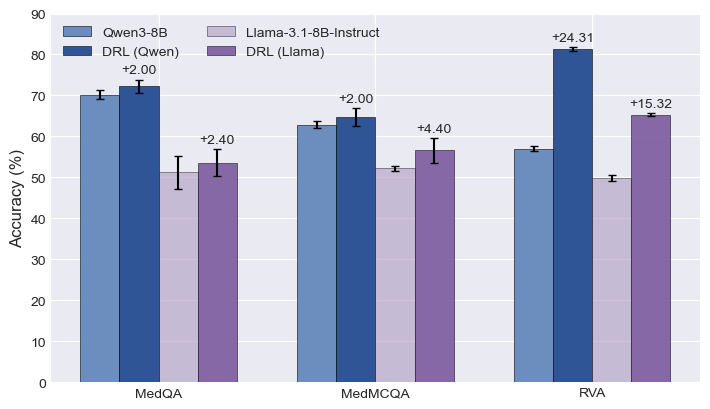}
  \caption{Performance gain of DRL over baselines across datasets.}
  \label{fig:performance_gain}
\end{figure}

Figure~\ref{fig:performance_gain} summarizes accuracy across the three datasets for DRL with two backbone models: Qwen and LLaMA. DRL consistently outperforms its corresponding base model, indicating that gains stem from the proposed discrepancy-driven augmentation rather than model-specific idiosyncrasies. Improvements on MedQA and MedMCQA are modest but consistent (roughly 2--4 points), whereas the largest gains are observed on RVA (approximately +24 points for Qwen and +15 points for LLaMA). This pattern suggests that DRL may be especially effective in the more reasoning-intensive clinical setting. Because RVA QA is derived from long, information-dense ED notes, and the question asks for prognosis, it imposes substantially higher reasoning demands than typical benchmark questions. Correspondingly, the base models perform comparably or worse on RVA than on MedQA/MedMCQA despite RVA being a binary prediction task. 

Table~\ref{tab:main_results} summarizes the comparison between DRL and other strong baseline models across datasets.
\begin{table}[htbp]
  \centering
  \small
  \caption{Main accuracy comparison results across datasets.\tablefootnote{DRL reports the best-performing top-$k$; RVA results use physician rationales.}}
  \begin{tabular}{lccc}
    \hline
    \textbf{Method} & \textbf{MedQA} & \textbf{MedMCQA} & \textbf{RVA} \\
    \hline
    DRL(Qwen) & 72.20$\pm$1.64 & 64.80$\pm$2.17 & \textbf{81.28$\pm$0.47} \\
    Qwen3-8B & 70.20$\pm$1.10 & 62.80$\pm$0.84 & 56.97$\pm$0.57 \\
    \midrule
    DRL(LLaMA) &  53.60$\pm$3.36 & 56.60$\pm$3.05 & 65.23$\pm$0.37 \\
    LLaMA-3.1-8B-Instruct & 51.20$\pm$4.09 & 52.20$\pm$0.58 & 49.91$\pm$0.69 \\
    \midrule
    MedReason-8B & \textbf{72.40$\pm$0.55} & 56.80$\pm$0.37 & 53.49$\pm$1.43 \\
    HuatuoGPT-o1-8B & 60.80$\pm$1.10 & \textbf{65.00$\pm$1.10} & 49.54$\pm$0.99 \\
    MedPRM-8B & 62.20$\pm$0.84 & 52.20$\pm$0.49 & 47.98$\pm$0.87 \\
    \hline
  \end{tabular}
  \label{tab:main_results}
\end{table}

\paragraph{Open medical QA benchmarks.}
With Qwen as the backbone, DRL achieves 72.20$\pm$1.64 on MedQA and 64.80$\pm$2.17 on MedMCQA. It consistently outperforms the base Qwen3-8B and all fine-tuned baselines, with the exception of HuatuoGPT-o1-8B on MedMCQA and MedReason-8B on MedQA. Notably, DRL performs on par with HuatuoGPT-o1-8B on MedMCQA and MedReason-8B on MedQA, despite being restricted to access only a small fraction of the training data. With LLaMA as the backbone, DRL similarly improves over LLaMA-3.1-8B-Instruct (53.60$\pm$3.36 vs. 51.20$\pm$4.09 on MedQA; 56.60$\pm$3.05 vs. 52.20$\pm$0.58 on MedMCQA), though it remains below specialized medical models, while avoiding the cost and reduced flexibility of fine-tuning. Overall, DRL yields reliable gains across backbones, whereas general-purpose instruction-tuned models without domain-specific reasoning support (e.g., LLaMA-3.1-8B-Instruct) perform substantially worse, underscoring the difficulty of these benchmarks without targeted reasoning guidance.

\paragraph{RVA-QA and domain shift.}
The largest gain is observed on RVA-QA: DRL(Qwen) achieves 81.28$\pm$0.47 accuracy, outperforming the strongest baseline (Qwen3-8B, 56.97$\pm$0.57) by 24.31 points with low variance. DRL(LLaMA) yields similar improvements over LLaMA-3.1-8B-Instruct (65.23$\pm$0.37 vs.\ 49.91$\pm$0.69; +15.32 points). In contrast, models fine-tuned on open medical QA (HuatuoGPT-o1-8B, MedReason-8B, MedPRM-8B) transfer poorly to RVA-QA, indicating limited robustness and generalizability under clinical task shift. These results suggest that discrepancy-mined, retrieval-based instructions more effectively address the missing factors and structural reasoning demands induced by long, information-dense clinical QA tasks.

\paragraph{Ablations: physician rationales, retrieval depth, and ICL}
To further validate and analyze our method's effectiveness, we conduct the following ablation studies on the RVA-QA dataset using Qwen as the base model:
\begin{enumerate}
  \item Include physician-authored clinical rationales in the prompt when constructing the reference reasoning graph $G_{ref}$ (results reported in Table~\ref{tab:main_results}).
  \item Exclude physician rationales from the prompt when constructing $G_{ref}$ to isolate their impact on performance.
  \item Compare against an in-context learning (ICL) baseline using the same training dataset to quantify the benefit of learning from reasoning differences versus learning from exemplars.
\end{enumerate}
\begin{table}[htbp]
  \centering
  \small
  \caption{Ablation on physician rationales and top-$k$ retrieval.\tablefootnote{ICL $k$=10 exceeds token limit}}
  \begin{tabular}{lcccc}
    \hline
     & Top-$k$=1 & Top-$k$=3 & Top-$k$=5 & Top-$k$=10 \\
    \hline
    With PHY. RAT. & 74.59$\pm$0.47 & 80.00$\pm$0.23 & 81.28$\pm$0.47 & 81.28$\pm$0.53 \\
    No PHY. RAT. & 72.75$\pm$0.43 & 80.37$\pm$0.27 & 79.63$\pm$0.34 & 80.92$\pm$0.47 \\
    ICL & 68.72$\pm$0.53 & 77.80$\pm$0.49 & 74.68$\pm$0.47 & - \\
    \hline
  \end{tabular}
  \label{tab:ablation_physician_reasoning}
\end{table}

%\begin{figure}[htbp]
%  \centering
%  \includegraphics[width=0.9\linewidth]{figs/top_k_comparison.png}
%  \caption{Impact of top-$k$ retrieval on alignment accuracy for different methods.}
%  \label{fig:top_k_vs_accuracy}
%\end{figure}

Table~\ref{tab:ablation_physician_reasoning} compares our method with and without physician rationales across different top-$k$ values, alongside the ICL baseline. ICL was not applicable at top-$k$=10 due to context length constraints. 
% Figure~\ref{fig:top_k_vs_accuracy} presents the corresponding visualization.
The results demonstrate that physician rationales consistently improve performance, especially at lower top-$k$ values, indicating that expert knowledge enhances reference reasoning graph quality and enables more effective learning from differential reasoning. This suggests that including physician rationale helps identify critical reasoning components the agent may overlook, thereby constructing more reliable near-gold-standard reasoning graphs that guide more accurate instruction generation. Furthermore, the ICL baseline consistently underperforms both variants of our method, demonstrating that DRL provides more targeted guidance than traditional in-context learning. This confirms that learning from reasoning graph differences under reference model supervision yields more effective clinical reasoning improvements than merely providing exemplar cases.

\paragraph{Effect of backbone strength.}
Table~\ref{tab:base_models_vs_accuracy} shows that Qwen3-8B is substantially stronger than LLaMA-3.1-8B-Instruct on RVA-QA across all Top-$k$ values, and DRL improves both backbones consistently. This suggests that stronger base reasoning capabilities and retrieval-based patching are complementary: higher-capacity backbones benefit from the same discrepancy-mined guidance, while weaker backbones still gain from targeted corrections.

% \begin{figure}[htbp]
%   \centering
%   \includegraphics[width=0.9\linewidth]{figs/base_models_comparison.png}
%   \caption{Impact of base models on alignment accuracy for different top-$k$ values.}
%   \label{fig:base_models_vs_accuracy}
% \end{figure}

\begin{table}[htbp]
  \centering
  \small
  \caption{Impact of base models on alignment accuracy for different top-$k$ values.}
  \begin{tabular}{lccccc}
    \hline
     & Top-$k$=0 & Top-$k$=1 & Top-$k$=3 & Top-$k$=5 & Top-$k$=10 \\
    \hline
    Qwen & 56.97$\pm$0.57 & 74.59$\pm$0.47 & 80.00$\pm$0.23 & 81.28$\pm$0.47 & 81.28$\pm$0.53 \\
    LLaMA & 49.91$\pm$0.69 & 62.29$\pm$0.79 & 64.13$\pm$0.39 & 61.65$\pm$0.11 & 65.23$\pm$0.37 \\
    \hline
  \end{tabular}
  \label{tab:base_models_vs_accuracy}
\end{table}

\paragraph{Physician qualitative review (RVA-QA)}
To complement automatic metrics with clinical face validity, three experienced clinicians reviewed reasoning outputs from both the base agent and DRL-corrected outputs for RVA-QA cases (see Appendix~\ref{sec:medical_scenarios} for complete case details). Clinicians were given the clinical vignette, the agent's answer and the ground truth (i.e., did an RVA actually occur), the agent reasoning, and DRL-generated assessments and instructions for improvement. We report summaries of three representative RVA cases to illustrate both the advantages and the limitations identified by clinician review.

\begin{itemize}
 \item \textbf{Urinary retention (agent: no RVA; ground-truth: no RVA):} The clinician noted key discharge cues (well-appearing or functional status; clear ED intervention such as Foley). They also agreed with the DRL assessment that the agent reasoning over-branched and referenced undocumented labs.

 \item \textbf{Breast rash/wound infection (agent: no RVA; ground-truth: RVA):} The clinician judged RVA risk as moderate to high due to baseline risk, leukocytosis, and concerning wound features, and emphasized that labs + explicit 24--48\,h return precautions signal ED concern. They endorsed DRL's assessment that the agent under-synthesized these risk factors; DRL guidance aligned with clinician feedback.

 \item \textbf{Fall + rib fracture + tube bleeding (agent: RVA; ground-truth: RVA):} The clinician noted a reasoning weakness in the agent: mixing evidence across problems and underweighting key admission cues. The DRL guidance was consistent with this feedback. The clinician also flagged an inconsistency in the DRL assessment, which stated the agent failed to predict disposition despite a correct prediction.
\end{itemize}

Overall, clinicians' reviews indicate that DRL-corrected reasoning better aligns with clinical judgment. In cases where the base agent reasoning exhibited deficiencies (speculative branching, evidence mismatches, under-weighting of key clinical signals), DRL instructions systematically corrected these gaps. However, in some cases, DRL failed to surface certain nuanced deficiencies that clinicians identified (e.g., subtle cues about treatment response likelihood). These findings suggest that while DRL substantially improves reasoning quality, further refinement of the base agent prompt, specifically, including relevant clinical assessments and grounding all reasoning strictly in documented evidence, would reduce the initial deficiency rate and enhance DRL's effectiveness.

\section{Conclusion}
We introduced \textbf{Differential Reasoning Learning (DRL)}, a reasoning graph discrepancy-driven framework for improving clinical agents beyond final-answer supervision. DRL contrasts reference rationales (from physicians, guidelines, or higher-capacity teachers) and an agent's CoT as typed reasoning DAGs, performs a clinically weighted GED-based discrepancy analysis with semantic node alignment, and distills recurring failure patterns into reusable natural-language patches stored in a Differential Reasoning Knowledge Base (DR-KB). At inference time, DRL retrieves and injects the most relevant patches to proactively close potential logic gaps. Across MedQA, MedMCQA, and an RVA-QA task, DRL is on par with baseline models for simple questions and outperforms baseline models for complex reasoning tasks derived from real-world care.

%\paragraph{Limitations and future directions.}
This paper has several limitations. First, DRL depends on LLM-based graph extraction and LLM-as-a-judge matching; extraction noise or adjudication bias can propagate into the mined patches, motivating stronger schema validation, uncertainty-aware discrepancy scoring, and selective human review for high-stakes cases. Second, DRL assumes access to high-quality reference rationales; thus, broadening coverage to settings with weaker supervision (e.g., noisy guidelines or incomplete physician rationales) remains important. Relatedly, we acknowledge that while gold standards such as guidelines exist, much of clinical medicine is considered more an art than a science and a gold standard is not available in all cases encountered in the real world clinical setting \cite{zhang2015paving}. Hence, triangulating between reference reasoning rationales tries to approximate evidence-based decision making with the best available knowledge. Third, while our evaluations had some clinicians' reviews, future work should include larger-scale expert assessment. Since each phase of knowledge mining is currently specific to the training cases, future work should also evaluate the scalability of DR-KB. Finally, future work can move from flat patch retrieval to extracting and validating conditional decision-rule fragments from discrepancy-context pairs, approximating an explicit latent decision policy that yields more reliable and auditable context augmentation.

\begin{acks}
The work of Ramayya Krishnan and Rema Padman was funded in part by the National Institute of Standards and Technology (NIST) through Federal Award ID Number 60NANB24D231 to the Carnegie
Mellon University AI Measurement Science and Engineering Center (AIMSEC). This research was also funded in part by the National Institute of Health (NIH) awards R01AG076998, R01AG089445, R01AG083039, and R01AI172875. We thank Drs. Justine Ko, Rochelle Joly, and Edward J. Schenck for providing clinical reviews of DRL outputs.
\end{acks}

\newpage

% \section{GenAI Disclosure}
% We partially relied on generative AI tools to support writing and editing. Specifically, we used ChatGPT 5.2 and Gemini 3.0 for sentence-level polishing and grammatical corrections, and used Gemini 3.0 (Nano Banana) to assist in generating Figure~\ref{fig:main_pipeline}. We also used AI-assisted coding tools (Codex, Claude Code and Cursor) to support our experiments.

\bibliographystyle{ACM-Reference-Format}
\bibliography{alignment}

\onecolumn
\appendix
\section{Prompts}
\subsection{Reasoning Graph extraction prompt for QA pairs}\label{sec:graph_extraction_prompt}
\begin{verbatim}
CLINICAL_NOTE_QA_PROMPT = '''
You are a medical AI expert. You will be give a question generated from clinical note, and corresponding answer.
Extract the reasoning graph from this question-answer pair.

## QUESTION generated from CLINICAL NOTE:
{question}
## ANSWER:
{answer}

## EXTRACTION INSTRUCTIONS:

1. **FACTS (F1, F2, ...)**: Extract observable clinical findings:
   - Vital signs, lab values, physical exam findings
   - Patient demographics, medical history
   - Symptoms reported, medications listed
   - Use "present", "absent", or "uncertain" for polarity
   - Include verbatim quote from the note as support

2. **HYPOTHESES (H1, H2, ...)**: Extract diagnoses or clinical conditions:
   - Primary and secondary diagnoses mentioned
   - Suspected or ruled-out conditions
   - Problem list items that represent clinical reasoning
   - Confidence based on how definitively stated in the note

3. **ACTIONS (A1, A2, ...)**: Extract clinical actions documented:
   - Medications prescribed (PRESCRIBE)
   - Tests ordered (TEST)
   - Treatments given (TREAT)
   - Assessments performed (ASSESS)
   - Monitoring plans (OBSERVE)

4. **FINAL**: The final answer of the question, which is given
   
5. **EDGES (E1, E2, ...)**: Infer connections between nodes:
   - Fact → Hypothesis: What findings support which diagnoses?
   - Hypothesis → Action: What conditions led to which treatments?
   - Fact → Action: Direct fact-to-action relationships
   - Action → Hypothesis: What actions suggest which diagnoses?
   - Fact or Hypothesis or Action → FINAL: What overall conclusion is supported?
   - Use "supports", "contradicts", or "suggests_test" as edge types
   - For each edge, there should be only one source and one destination


## OUTPUT SCHEMA:
{{
  "artifact_id": "{artifact_id}",
  "qid": "{qid}",
  "guideline_bundle_id": "extracted",
  "facts": [
    {{
      "fid": "F1",
      "label": "finding description",
      "value": "specific value or null",
      "polarity": "present|absent|uncertain",
      "support": [{{"source": "case", "ref": null, "quote": "verbatim from note"}}]
    }}
  ],
  "hypotheses": [
    {{
      "hid": "H1",
      "label": "diagnosis or condition",
      "confidence": "high|medium|low|uncertain",
      "support": [{{"source": "case", "ref": "F1", "quote": "supporting text"}}]
    }}
  ],
  "actions": [
    {{
      "aid": "A1",
      "label": "action description",
      "action_type": "TEST|TREAT|ASSESS|OBSERVE|PRESCRIBE",
      "status": "recommended|contraindicated|optional|not_specified",
      "support": [{{"source": "case", "ref": null, "quote": "from note"}}]
    }}
  ],
  "edges": [
    {{
      "eid": "E1",
      "src": "F1",
      "dst": "H1",
      "type": "supports|contradicts|suggests_test",
      "justification": "why this connection exists based on the note",
      "guideline_support": null
    }}
  ],
  "final": {{
    "answer": "{answer}",
    "answer_text": "reasoning behind the answer",
    "confidence": "high|medium|low|uncertain",
    "uncertainties": ["list of unclear aspects"]
  }}
}}

Return ONLY the JSON object.
'''
\end{verbatim}

\subsection{GED and instruction prompt}\label{sec:ged_instruction_prompt}
\begin{verbatim}
prompt = """You are a medical expert evaluating an agent's reasoning quality using Graph Edit Distance (GED).

QUESTION: {question}
CORRECT ANSWER: {answer}
AGENT ANSWER CORRECT: {'YES' if answer_correct else 'NO'}

REFERENCE GRAPH (G_REF):
Facts: {g_ref_facts}
Hypotheses: {g_ref_hyps}
Actions: {g_ref_actions}

AGENT GRAPH (G_AGENT):
Facts: {g_agent_facts}
Hypotheses: {g_agent_hyps}
Actions: {g_agent_actions}

---

STEP 1: CALCULATE GED SCORE

Evaluate GED between G_REF and G_AGENT using these rules:

1. SEMANTIC NODE MATCHING: Match by meaning, not exact text
2. CONTEXT-BASED REASONABLENESS: Accept facts from question context
3. NODE IMPORTANCE: Facts=1.0x, Hypotheses=1.5x, Actions=2.0x
4. ANSWER IMPACT: 
   - Correct + correct path: GED 0.1-0.3
   - Correct + wrong path: GED 0.6-0.8
   - Wrong: GED > 0.8
5. REASONING PATH QUALITY: Check edge connectivity

IMPORTANT CALCULATION FORMULA:
- Calculate three penalty components (each between 0.0-1.0):
  * node_mismatch_penalty: Penalty for missing or mismatched nodes
  * hallucination_penalty: Penalty for hallucinated (extra) nodes
  * reasoning_path_penalty: Penalty for incorrect reasoning path structure
- Calculate total_ged_score EXACTLY as:
  total_ged_score = node_mismatch_penalty + hallucination_penalty + reasoning_path_penalty
- The total_ged_score can exceed 1.0 if the sum of penalties is greater than 1.0

---

STEP 2: GENERATE ENHANCED IMPROVEMENT INSTRUCTIONS

Generate improvement instructions that could be used in a future prompt 
when the agent encounters similar medical questions. These instructions should help 
the agent avoid the same mistakes and reason more effectively.

Each instruction should include:

{
  "title": "Brief error title",
  "error_type": "diagnosis_error|treatment_error|reasoning_error|missed_fact",
  "clinical_domain": "<domain>",
  
  "situation_context": "This instruction applies when: [clinical scenario where error occurred]",
  "example_situation": "For example: [specific example from this case]",
  
  "similar_case_patterns": [
    "Pattern 1: [how to recognize similar cases]",
    "Pattern 2: [clinical signs to watch for]",
    "Pattern 3: [key indicators]"
  ],
  
  "what_went_wrong": "[Specific mistake made]",
  "why_it_matters": "[Clinical consequences and impact]",
  
  "potential_risks": [
    "Risk 1: [specific bad outcome if error repeated]",
    "Risk 2: [patient safety impact]",
    "Risk 3: [clinical consequences]"
  ],
  
  "prevention_steps": [
    "Step 1: [concrete action]",
    "Step 2: [concrete action]",
    "Step 3: [concrete action]"
  ],
  
  "contraindications": [
    "DO NOT: [specific thing to avoid]",
    "NEVER: [explicit warning]",
    "AVOID: [specific practice]"
  ],
  
  "trigger_keywords": ["keyword1", "keyword2"],
  "medical_principle": "[Clinical foundation]",
  "similar_case_examples": "[When this guidance applies]"
}

Important: Include practical, actionable guidance that prevents the EXACT error that occurred.

---

CRITICAL: When calculating total_ged_score, you MUST:
1. First calculate node_mismatch_penalty, hallucination_penalty, and reasoning_path_penalty (each 0.0-1.0)
2. Then calculate: total_ged_score = node_mismatch_penalty + hallucination_penalty + reasoning_path_penalty
3. Do NOT use any other formula or normalization - use the exact sum of the three penalties

---

OUTPUT FORMAT:

Return ONLY valid JSON with these fields:

{{
  "qid": "{qid}",
  "answer_correct": {str(answer_correct).lower()},
  
  "node_matching": {{
    "matched_facts": <number>,
    "matched_hypotheses": <number>,
    "matched_actions": <number>,
    "hallucinated_nodes": <list or empty>,
    "hallucination_severity": "none|minor|moderate|severe"
  }},
  
  "graph_structure": {{
    "reasoning_path_correct": <boolean>,
    "missing_edges": <list>,
    "extra_edges": <list or empty>
  }},
  
  "ged_calculation": {{
    "node_mismatch_penalty": <0.0-1.0>,
    "hallucination_penalty": <0.0-1.0>,
    "reasoning_path_penalty": <0.0-1.0>,
    "total_ged_score": <EXACTLY: node_mismatch_penalty + hallucination_penalty + reasoning_path_penalty>,
    "explanation": "<brief explanation>"
  }},
  
  "assessment": {{
    "overall": "Excellent|Good|Fair|Poor|Very Poor",
    "strengths": <list of 1-2 strengths>,
    "weaknesses": <list of 1-2 weaknesses>
  }},
  
  "improvement_instructions": [
    {{
      "title": "Brief title",
      "error_type": "diagnosis_error|treatment_error|reasoning_error|missed_fact",
      "clinical_domain": "<domain>",
      "situation_context": "This instruction applies when: ...",
      "example_situation": "For example: ...",
      "similar_case_patterns": ["Pattern 1: ...", "Pattern 2: ..."],
      "what_went_wrong": "...",
      "why_it_matters": "...",
      "potential_risks": ["Risk 1: ...", "Risk 2: ...", "Risk 3: ..."],
      "prevention_steps": ["Step 1: ...", "Step 2: ...", "Step 3: ..."],
      "contraindications": ["DO NOT: ...", "NEVER: ...", "AVOID: ..."],
      "trigger_keywords": ["keyword1", "keyword2"],
      "medical_principle": "...",
      "similar_case_examples": "..."
    }}
  ]
}}

Return ONLY JSON, no other text."""
\end{verbatim}

\subsection{Prompt for converting clinical notes into QA pairs}\label{sec:clinical_note_to_qa_prompt}
\begin{verbatim}
  prompt = """
You are a clinician-grade medical language model. Your task is to transform emergency department (ED) clinical notes into a single, comprehensive clinical question suitable for clinical prognosis and reasoning. 
You are given an emergency department clinical note composed of the following sections:
- Chief Complaint
- Patient History
- Review of Systems
- Physical Exam
- Medical Decision Making
- Assessment and Plan
Rewrite the note as one self-contained clinical question that could be posed to a clinical decision-support system or clinician at the time of ED discharge. The clinical question is concerned with whether this patient will return to ED and get admitted to inpatient within 9 days after the discharge of the index ED visit. The answer should be yes or no. 
Constraints:
1. Do not omit any clinical facts, findings, symptoms, comorbidities, diagnostics, treatments, or disposition details.
2. Do not introduce new information or reinterpret findings.
3. Preserve uncertainty, negations, and clinician reasoning exactly as stated.
4. Use medically precise language.
5. Do not answer the question.
6. Do not include probabilities, risk estimates, or predictions.
7. Remove all names of patients and physicians and all other identifying information.
Output format:
1. A single paragraph written as a clinical question beginning with:“Given the following emergency department presentation and evaluation…”
2. The question must end with: “Will this patient return to ED and get admitted to inpatient within 9 days after the discharge of the index ED visit”

Clinical Note:
{clinical_note}  
"""
\end{verbatim}

\section{Physician review details}\label{sec:medical_scenarios}

\subsection{Medical scenario 1 (de-identified)}

\begin{quote}
\small
\noindent\textbf{[\ldots]}\\
\textit{Medical background information omitted due HPI data.}\\
\textbf{[\ldots]}
\end{quote}
\vspace{0.5em}

\noindent\textbf{Question:} Will this patient return to ED and get admitted to inpatient within 9 days after the discharge of the index ED visit?

\noindent\textbf{True answer:} No

\noindent\textbf{Agent answer:} No

\noindent\textbf{Agent reasoning:}

\begin{quote}
\small
\noindent The patient is an 84-year-old male with a history of BPH and HLD, but the medical history isn't fully documented. He's presenting with urinary retention, decreased output, dysuria, and intermittent hematuria. His son reported some fever and chills a week ago. The physical exam shows suprapubic tenderness, and the ED doctor suspects acute urinary retention, UTI, or cystitis. They plan a Foley catheter, antibiotics, and discharge.\par\medskip The question is whether he'll return to ED and be admitted within 9 days. The answer choices are 0 (no) or 1 (yes). \par\medskip First, I need to consider the possible diagnoses. The main issues here are urinary retention and possible UTI. BPH is a known cause of urinary retention, especially in older men. The suprapubic tenderness suggests bladder distension, which aligns with retention. The hematuria and dysuria could indicate a UTI or even a more severe issue like a bladder infection or stones. The fever and chills a week ago might have been a past infection, but the current symptoms are more acute.\par\medskip The ED plan includes a Foley catheter to relieve retention, which should address the immediate problem. Antibiotics are being considered, which is good for a UTI. However, the patient's history of BPH might mean that the retention is due to an obstructive issue, which could be managed with catheterization and possibly further urological evaluation. \par\medskip But wait, the differential includes possible cystitis and UTI. If the UTI is severe, it could lead to complications like pyelonephritis, but the ED doctor has low suspicion for that. The patient's vital signs are normal, and he's not in acute distress. The plan is for discharge, so they might think the issue is manageable outpatient. \par\medskip However, the presence of hematuria and the history of fever and chills might indicate a more serious infection. If the UTI is not adequately treated, it could lead to sepsis or kidney involvement. But the ED is planning antibiotics, which should cover that. Also, the Foley catheter would help with the retention, which might resolve the symptoms. \par\medskip Another angle: the patient's age and BPH increase the risk of complications. If the retention is due to BPH, then after catheterization, he might need further management like medication or surgery. But the question is about returning to ED and being admitted. If the initial treatment is effective, maybe he won't need to come back. However, if there's a complication like sepsis, AKI, or if the UTI is not responding to antibiotics, he might need readmission. \par\medskip The ED doctor's differential doesn't include high suspicion for pyelonephritis or AKI, so maybe they think it's a straightforward UTI or BPH-related retention. The planned evaluations are CBC, chem, UA, and UCx. If those results are normal, then discharge is appropriate. But if there's a complication, like a high WBC count or signs of infection, they might admit. \par\medskip But the question is about the likelihood of return within 9 days. If the patient is treated with a Foley and antibiotics, and the underlying cause (BPH or UTI) is managed, he might not need to come back. However, if the UTI is severe or if there's a complication, he might. But the ED's assessment is that the most likely is acute urinary retention, which is being treated. The answer choices are 0 (no) or 1 (yes). \par\medskip Considering the information given, the ED is planning for discharge, and the differential doesn't suggest high suspicion for severe complications. So the answer might be 0 (no). But I need to be careful. The presence of hematuria and fever could indicate a more serious issue. However, the ED doctor is considering UTI and cystitis, which are treatable. If the patient's symptoms resolve with treatment, he might not return. But if there's a complication, he might. However, the question is about the likelihood. Since the ED is planning for discharge and the differential doesn't include high suspicion for severe issues, the answer is probably 0.\newline </think>\par\medskip 1. The patient presents with acute urinary retention, dysuria, hematuria, and a history of BPH, which are classic signs of urinary retention or a urinary tract infection (UTI). \newline 2. The ED team has initiated a Foley catheter and planned antibiotics, addressing the immediate issue of retention and potential infection. \newline 3. The absence of severe symptoms like flank pain, high fever, or signs of sepsis, along with normal vital signs, suggests a less severe condition. \newline 4. The differential diagnosis includes UTI and BPH-related retention, both of which are typically managed with outpatient care, including antibiotics and urological follow-up. \newline 5. The low suspicion for pyelonephritis or acute kidney injury (AKI) further supports the likelihood of a manageable outpatient course. \newline 6. Without evidence of complications or treatment failure, the patient is likely to be discharged and not require readmission within 9 days. The answer is no.
\end{quote}

\noindent\textbf{DRL-generated assessment on agent reasoning:}

\begin{center}
\begingroup
\small
\setlength{\tabcolsep}{4pt}
\renewcommand{\arraystretch}{1.15}
\begin{tabular}{@{}p{0.18\linewidth}p{0.78\linewidth}@{}}
\textbf{Overall} & Fair\\
\textbf{Strength} & Correctly identified the primary diagnoses (urinary retention, UTI, BPH) and key symptoms\newline Correctly identified the immediate ED plan (Foley, antibiotics, discharge) and reached the correct final answer\\
\textbf{weaknesses} & Significant number of hallucinated hypotheses and actions, some of which contradict the case (e.g., 'admit' vs. 'anticipated discharge')\newline Missed several specific facts from the ROS and physical exam, and miscategorized 'discharge' from a fact to an action\\
\textbf{generated instruction} & These guidelines are crucial for enhancing medical reasoning, particularly in Emergency Medicine, by addressing common errors in interpreting clinical information. Firstly, avoid hallucinating speculative diagnoses or hypotheses; generating broad, unsupported diagnoses not explicitly considered by the ED physician leads to unfocused and misleading differentials, obscuring the actual assessment and suggesting a lack of contextual understanding. This error often occurs when agents list potential diagnoses beyond the immediate ED scope or not directly supported by clinical findings. To prevent this, strictly adhere to the physician's stated differential and level of suspicion, inferring only directly supported hypotheses, and differentiating between comprehensive textbook knowledge and the specific differential being considered, especially when trigger keywords like "low suspicion for" or "assessed as" are present. Secondly, it is imperative to adhere to the physician's stated plan and avoid contradictory actions. Generating actions not part of the immediate ED plan, or directly contradicting the stated disposition (e.g., listing 'admit' when 'discharge' is planned), misrepresents the physician's decision-making and care pathway. This includes inferring long-term management not relevant to the acute ED visit. Prevention involves prioritizing the physician's explicit 'Evaluations planned,' 'Treatments,' and 'Disposition,' ensuring consistency, and avoiding speculation on future treatments beyond the current encounter, particularly when phrases like "Disposition is anticipated" are used. Finally, maintain node type fidelity by accurately categorizing information as Facts, Hypotheses, or Actions. Miscategorizing a descriptive statement about a plan or outcome (like 'Disposition anticipated discharge' being a Fact) as an active intervention (an Action) distorts the graph's structure and misrepresents the clinical narrative. To prevent this, carefully distinguish between descriptive statements, diagnostic considerations, and planned interventions, understanding that a 'disposition' is a plan/outcome statement (a fact), not an action, especially when trigger keywords like "plan is" or "reveals" are encountered. In all cases, avoid adding unmentioned or unimplied information, never include contradictory or highly speculative elements, and always ensure precise categorization to reflect the physician's immediate assessment and plan.,\\
\end{tabular}
\endgroup
\end{center}

\noindent\textbf{Clinician's review: Urinary retention (Agent predicted no RVA; clinician agrees)}

The clinician emphasized that bedside appearance and functional status (e.g., well-appearing, ambulatory, tolerating PO) and a clear ED intervention (Foley catheter) are key signals supporting discharge. The base agent reasoning occasionally over-elaborated counterfactual admit/discharge branches and referenced undocumented laboratory results. DRL instructions explicitly corrected these issues by: (i) anchoring reasoning on documented findings only, and (ii) prioritizing functional status and direct interventions. The clinician agreed with DRL-generated assessment that the agent gives multiple possible admit/discharge branches, indicating these instructions are reasonable.

\subsection{Medical scenario 2 (de-identified)}

\begin{quote}
\small
\noindent\textbf{[\ldots]}\\
\textit{Medical background information omitted due HPI data.}\\
\textbf{[\ldots]}
\end{quote}
\vspace{0.5em}

\noindent\textbf{Question:} Will this patient return to ED and get admitted to inpatient within 9 days after the discharge of the index ED visit?

\noindent\textbf{True answer:} Yes

\noindent\textbf{Agent answer:} No (wrong answer)

\noindent\textbf{Agent reasoning:}

\begin{quote}
\small
\noindent The patient is a 65-year-old female (though referred to with female pronouns, but the medical history mentions male with ESI 4, which might be a typo or confusion. Wait, the initial note says "male with an ESI of 4" but then refers to her with female pronouns. Maybe that's a mistake, but I'll proceed with the info given.\par\medskip She has a rash on inner thighs and left breast, described as raw, weeping dark fluid with a foul smell. Physical exam shows erythematous, tender skin with satellite lesions, and some raw areas. Lab shows WBC 14.12, which is elevated. She was treated for intertrigo with nystatin and Keflex. She was discharged with instructions to return in 24-48 hours or sooner if worsening, and to see dermatology.\par\medskip The question is whether she'll return to ED and get admitted within 9 days. The answer choices are 0 (no) or 1 (yes).\par\medskip First, consider the initial diagnosis: intertrigo. But the presentation includes foul-smelling discharge, which might suggest a fungal infection (like Candida) or bacterial infection. The left breast has an ulcer with fat layer exposed and cellulitis. The groin and thighs have superficial ulcerations with foul smell and satellite lesions, suggesting a fungal infection with possible secondary bacterial infection.\par\medskip The patient has some comorbidities listed (asthma, beta thalassemia, cyclic neutropenia, sarcoidosis), and diabetes is in the PMH but she denies it. Her glucose is normal. But cyclic neutropenia could predispose to infections.\par\medskip She is treated with antibiotics (Keflex) and nystatin, and referred to dermatology. She is discharged stable, ambulatory.\par\medskip The question is about return within 9 days and admission. Given that she has significant skin infection and ulceration, and elevated WBC, there is a risk of worsening infection, sepsis, or needing IV antibiotics. But they treated with oral antibiotics and antifungal, and she's stable.\par\medskip However, the ulcer of breast with fat layer exposed is concerning. Also, the foul smell suggests possible necrotizing infection? But no mention of systemic symptoms or high fever. Her vitals are normal and she is not ill-appearing.\par\medskip She is instructed to return in 24-48 hours for wound check. That suggests the ED expects possible progression or need for further care. But that could be outpatient wound check, not necessarily admission. But if she returns, she might not be admitted.\par\medskip Still, the elevated WBC and underlying neutropenia and sarcoidosis could mean she is immunocompromised, which increases risk.\par\medskip But the question is admission within 9 days, not just return. Many intertrigo cases can be managed outpatient with topical and oral meds.\par\medskip If she returns, it might be for worsening cellulitis or failure of oral antibiotics. But admission is possible if infection worsens.\par\medskip But given she is stable, no fever, and treated, I lean towards no admission.\par\medskip So answer likely 0.\newline </think>\par\medskip 1. Patient has intertrigo with ulceration and cellulitis, treated with appropriate outpatient medications (nystatin, Keflex).\newline 2. Vital signs stable, no systemic symptoms, and patient ambulatory and not ill-appearing.\newline 3. Although instructed to return for wound check, this is likely precautionary and for outpatient follow-up.\newline 4. Without evidence of treatment failure, severe systemic infection, or inability to manage at home, inpatient admission is unlikely within 9 days.\newline 5. Therefore, the patient is not expected to return and be admitted.\newline \newline <answer>0</answer>
\end{quote}

\noindent\textbf{DRL-generated assessment on agent reasoning:}

\begin{center}
\begingroup
\small
\setlength{\tabcolsep}{4pt}
\renewcommand{\arraystretch}{1.15}
\begin{tabular}{@{}p{0.18\linewidth}p{0.78\linewidth}@{}}
\textbf{Overall} & Very Poor\\
\textbf{Strength} & "Identified the primary diagnosis (intertrigo) and key treatments.",\newline Recognized the penicillin allergy and the rationale for Keflex use."\\
\textbf{weaknesses} & "Failed to synthesize critical risk factors (elevated WBC, comorbidities, severity of rash, explicit return instructions) to predict the likelihood of readmission/admission.",\newline "Hallucinated outcomes and predictions as 'facts' or 'actions', which directly led to the incorrect answer and skewed its reasoning."\\
\textbf{generated instruction} & These instructions highlight critical reasoning errors in medical practice, emphasizing the need for precise information processing. First, when evaluating patients for ED discharge, especially those with infections or complex medical histories, it is crucial to integrate a multi-factor risk assessment to accurately predict readmission or inpatient admission. The common error is failing to synthesize cumulative risks from factors like age, comorbidities, objective infection severity (e.g., elevated WBC, foul smell, raw skin, difficulty walking), and explicit discharge instructions (e.g., "return to ED for wound check"), which leads to underestimating risk, potential patient harm, delayed care, and increased costs. To prevent this, systematically review all relevant patient data, identify 'red flag' indicators, and pay close attention to discharge instructions suggesting early follow-up, rather than solely relying on a patient's subjective 'feeling well' or discharge status. Second, avoid hallucinating or misclassifying outcomes as facts or actions in clinical reasoning. This error occurs when clinicians or models assume a patient's future course (e.g., "will likely not return") as an established fact or treat it as a planned action, leading to flawed decision-making. Prevention requires strict adherence to the presented clinical facts and physician's explicit plan, ensuring that predictions remain separate from documented events or intended interventions. When extracting or generating clinical reasoning graphs, categorize information correctly: facts are objective and documented, hypotheses are diagnostic considerations, actions are planned interventions, and outcomes are distinct future events that must not be assumed unless explicitly stated. Always avoid adding unmentioned details, ensure consistent categorization, and prioritize cautious, evidence-based risk synthesis for patient safety.
\end{tabular}
\endgroup
\end{center}

\noindent\textbf{Clinician's review: Rash / wound infection under breast (Agent predicted no RVA; clinician disagreed)}

The clinician assessed return risk as moderate to high given baseline risk factors (diabetes; sarcoidosis/possible immunosuppression; history of neutropenia), leukocytosis (WBC $14.12$), and concerning wound features (foul odor, yellow drainage, multifocal involvement). The clinician further noted that obtaining labs for a wound infection and providing explicit 24--48\,h return precautions are strong signals of clinician concern. The base agent oscillated between low and high risk without anchoring on baseline vulnerability or likelihood of treatment failure. DRL instructions corrected this by: (i) explicitly weighting baseline comorbidities and immunosuppression, (ii) flagging when lab workup indicates higher concern, and (iii) recognizing explicit return precautions as high-risk signals. These corrections directly addressed the reasoning gaps the clinician identified. The weaknesses list generated by DRL also matched what the clinician said: elevated WBC, comorbidities, rash appearance, explicit return precautions.

\subsection{Medical scenario 3 (de-identified)}

\begin{quote}
\small
\noindent\textbf{[\ldots]}\\
\textit{Medical background information omitted due HPI data.}\\
\textbf{[\ldots]}
\end{quote}
\vspace{0.5em}

\noindent\textbf{Question:} Will this patient return to ED and get admitted to inpatient within 9 days after the discharge of the index ED visit?

\noindent\textbf{True answer:} Yes

\noindent\textbf{Agent answer:} Yes (correct answer)

\noindent\textbf{Agent reasoning:}

\begin{quote}
\small
\noindent The patient is an 81-year-old male with a complex medical history. He presented to the ED with a fall and difficulty breathing. The main issues are left-sided rib and back pain, along with pink-tinged urine from his left PCNU. He also has a history of CKD stage 4, heart failure, and multiple other conditions.\par\medskip First, I need to consider the possible reasons for his symptoms. The chief complaints are a fall and difficulty breathing. The fall caused left-sided chest trauma, leading to rib and back pain. The PCNU site has pink-tinged urine and surrounding erythema, which might indicate a complication like infection or bleeding. His weight loss and decreased oral intake could contribute to weakness and possibly malnutrition, which might affect his recovery.\par\medskip The differential diagnosis includes rib fracture, musculoskeletal injury, PCNU complications (like infection or blockage), and maybe even a urinary tract infection (UTI) or pyelonephritis. The CXR didn't show pneumonia, so that's less likely. His shortness of breath could be due to COPD exacerbation or heart failure exacerbation, especially given his low EF and recent admissions.\par\medskip The ED attending anticipates admission. The patient is ill-appearing, cachectic, tachycardic, and has multiple comorbidities. He has had multiple admissions in the last month and was discharged yesterday. This suggests he is unstable and likely to require inpatient care.\par\medskip Additionally, the fact that he's too weak to cook and has decreased oral intake suggests functional decline. His fall and pain could limit mobility. His shortness of breath at rest for days suggests a significant pulmonary or cardiac issue. Also, the nephrostomy tube complications could indicate infection, which in a frail patient could require admission.\par\medskip Even though the chest X-ray didn't show pneumonia, he could still have CHF exacerbation or COPD exacerbation. The ED planned labs including BNP and troponin, which suggests concern for cardiac issues.\par\medskip Given all these factors, it's highly likely that he will be admitted during this visit. The question asks whether he will return to ED and be admitted within 9 days after discharge of the index ED visit. But note, the index visit is this one, and disposition is anticipated admission. Wait, the question: "return to ED and get admitted within 9 days after discharge of the index ED visit." If he is admitted now, then he hasn't been discharged yet. But perhaps the index ED visit ends with discharge, and then we look at return visits. Hmm. Actually, in the prompt, the disposition is anticipated admission, but maybe he was discharged? It's confusing. However, since the true answer is yes and AI answer is yes, likely he was discharged and then returned.\par\medskip But the ED plan anticipates admission, so he likely will be admitted. The patient has multiple risk factors for readmission, including chronic illness, recent admissions, functional decline, and device complications. So yes, he is likely to return and be admitted.\newline </think>\par\medskip 1. The patient is an 81-year-old male with multiple chronic comorbidities (CKD stage 4, HFrEF with EF 15-20\%, CAD, COPD, cachexia) and has had multiple recent admissions, including discharge just yesterday.\newline 2. He presents with significant acute issues: a fall with severe left-sided rib pain, shortness of breath at rest for several days, tachycardia, and signs of nephrostomy tube complications (pink output, erythema), all of which increase the likelihood of clinical deterioration.\newline 3. He is ill-appearing, cachectic, and functionally declining (decreased oral intake, too weak to cook), suggesting inability to safely manage at home.\newline 4. The ED clinician's assessment includes concerns for serious conditions (rib fracture, COPD/CHF exacerbation, nephrostomy infection) and anticipates admission.\newline 5. Given his frailty, complexity, and acute presentation, the risk of readmission/inpatient admission within 9 days is high.\newline \newline <answer>1</answer>
\end{quote}

\noindent\textbf{DRL-generated assessment on agent reasoning:}

\begin{center}
\begingroup
\small
\setlength{\tabcolsep}{4pt}
\renewcommand{\arraystretch}{1.15}
\begin{tabular}{@{}p{0.18\linewidth}p{0.78\linewidth}@{}}
\textbf{Overall} & Very Poor\\
\textbf{Strength} & "Identified some key patient facts and the primary diagnostic hypotheses considered by the clinician."\\
\textbf{weaknesses} & "Failed to correctly predict the patient's disposition (admission/readmission).",\newline "Significant hallucination of general hypotheses and miscategorization of outcomes as 'actions'.",\newline Missed a large number of specific facts from the detailed patient presentation.,\newline Demonstrated severe internal inconsistency by listing 'admission' as an action while concluding 'NO' to admission."\\
\textbf{generated instruction} & This instruction addresses two critical reasoning errors: inaccurate risk assessment for admission/readmission in complex patients and inaccurate node categorization with excessive hallucination. When evaluating complex, elderly patients with multiple comorbidities presenting with new symptoms (e.g., falls, weight loss, device issues, tachycardia), avoid underestimating cumulative risk factors, as this reasoning error can lead to inappropriate discharge, patient harm, and readmissions. Instead, synthesize all significant risk factors, consider their cumulative burden, prioritize patient safety by erring on the side of caution, and recognize red flags like tachycardia or new device-related symptoms. Simultaneously, when extracting and categorizing clinical information into Facts, Hypotheses, and Actions, prevent the reasoning error of miscategorization and vague hypotheses by strictly adhering to definitions: Facts are objective patient data, Hypotheses are diagnostic considerations, Actions are planned clinician interventions. Never include predicted outcomes as Actions, and avoid generating speculative hypotheses not grounded in the physician's stated differential. This ensures accurate clinical reasoning representation and avoids misleading graphs or downstream decisions.\\
\end{tabular}
\endgroup
\end{center}

\noindent\textbf{Clinician's review: Fall + rib fracture + tube bleeding (Agent predicted RVA; clinician agreed, but flagged reasoning)}

Although the final prediction matched, the clinician noted the base agent sometimes conflated evidence across concurrent problems (e.g., reasoning that a normal chest X-ray addresses renal/tube issues when these require different diagnostic modalities). The clinician also highlighted high-salience admission cues that the base agent underweighted, including ill-appearing/cachectic presentation and functional decline (reduced intake; ``too weak to cook''). DRL instructions addressed these by: (i) delineating reasoning for concurrent complaints separately, and (ii) elevating functional status and appearance as primary disposition signals. The clinician pointed out that the issue in the DRL-generated assessment that the agent failed to predict disposition even though it reasoned correctly.

\end{document}